\documentclass[manuscript,screen,review=false]{acmart}

\usepackage{graphicx}
\usepackage{algorithm}
\usepackage{algorithmicx}
\usepackage{algpseudocode}
\usepackage{amsfonts}
\usepackage{mathtools}
\usepackage{bm}
\usepackage{bbm}
\usepackage{multirow}
\usepackage{multicol}
\usepackage{abstract}
\usepackage{pifont}
\usepackage{xspace}
\usepackage{cleveref}

\AtBeginDocument{%
  \providecommand\BibTeX{{%
    \normalfont B\kern-0.5em{\scshape i\kern-0.25em b}\kern-0.8em\TeX}}}

\makeatletter
\DeclareRobustCommand\onedot{\futurelet\@let@token\@onedot}
\def\@onedot{\ifx\@let@token.\else.\null\fi\xspace}

\newcommand{\etal}{\emph{et~al\@\onedot}}

\newcommand{\ordinal}[1]{{#1}\textsuperscript{th}}
\newcommand{\method}{Fed-Credit}

\DeclarePairedDelimiter{\parens}{\lparen}{\rparen}
\DeclarePairedDelimiter{\angles}{\langle}{\rangle}
\DeclarePairedDelimiter{\bracks}{\lbrack}{\rbrack}
\DeclarePairedDelimiter{\braces}{\lbrace}{\rbrace}

\DeclarePairedDelimiter{\norm}{\lVert}{\rVert}

\newcommand{\lr}{\eta}
\newcommand{\rounds}{R}
\newcommand{\globalmodel}{{\bm\theta}}
\newcommand{\localmodel}{{\bm\theta_{i}}}

\newcommand{\credvals}{{\bm\tau}}
\newcommand{\credval}{{\tau_{i}}}
\newcommand{\credscore}{S_{i}}

\begin{document}

\title{\method{}: Robust Federated Learning with Credibility Management}

\author{Jiayan Chen}
\authornote{Both authors contributed equally to this research.}
\email{ryan-utopia@outlook.com}
\affiliation{%
    \institution{Beijing Normal University at Zhuhai}
    \streetaddress{No.18, Jinfeng Road}
    \city{Zhuhai}
    \state{Guangdong}
    \country{China}
    \postcode{519087}
}

\author{Zhirong Qian}
\authornotemark[1]
\email{zr.qian@outlook.com}
\affiliation{%
    \institution{Beijing Normal University at Zhuhai}
    \city{Zhuhai}
    \state{Guangdong}
    \country{China}
    \postcode{519087}
}

\author{Tianhui Meng}
\authornote{Corresponding author.}
\email{tmeng@bnu.edu.cn}
\affiliation{%
    \institution{Beijing Normal University at Zhuhai}
    \city{Zhuhai}
    \state{Guangdong}
    \country{China}
    \postcode{519087}
}
\author{Xitong Gao}
\email{xt.gao@siat.ac.cn}
\affiliation{%
    \institution{%
        Shenzhen Institute of Advanced Technology,
        Chinese Academy of Sciences}
    \city{Shenzhen}
    \state{Guangdong}
    \country{China}
}

\author{Tian Wang}
\email{tianwang@bnu.edu.cn}
\affiliation{%
    \institution{Beijing Normal University at Zhuhai}
    \city{Zhuhai}
    \state{Guangdong}
    \country{China}
    \postcode{519087}
}
\author{Weijia Jia}
\email{jiawj@bnu.edu.cn}
\affiliation{%
    \institution{Beijing Normal University at Zhuhai and Guangdong Key Lab of AI and Multi-Modal Data Processing, BNU-HKBU United International College and Beijing Normal University at Zhuhai}
    \city{Zhuhai}
    \state{Guangdong}
    \country{China}
    \postcode{519087}
}
\renewcommand{\shortauthors}{J. Chen, Z. Qian, T. Meng, X. Gao, T. Wang, W. Jia}

\keywords{Privacy preservation, Federated learning, Credibility management, Poisoning attacks}

\maketitle
\begin{abstract}
Aiming at privacy preservation,
Federated Learning (FL)
is an emerging machine learning approach
enabling model training on decentralized devices or data sources.
The learning mechanism of FL
relies on aggregating parameter updates
from individual clients.
However,
this process may pose a potential security risk
due to the presence of malicious devices.
Existing solutions are either costly
due to the use of compute-intensive technology,
or restrictive for reasons of strong assumptions
such as the prior knowledge of
the number of attackers
and how they attack.
Few methods consider both privacy constraints
and uncertain attack scenarios.
In this paper,
we propose a robust FL approach
based on the credibility management scheme,
called \method{}.
Unlike previous studies,
our approach does not require prior knowledge
of the nodes and the data distribution.
It maintains and employs a credibility set,
which weighs the historical clients' contributions
based on the similarity between the local models and global model,
to adjust the global model update.
The subtlety of \method{}
is that the time decay
and attitudinal value factor
are incorporated into the dynamic adjustment
of the reputation weights
and it boasts a computational complexity of \( O(n) \)
(\( n \) is the number of the clients).
We conducted extensive experiments
on the MNIST and CIFAR-10 datasets
under 5 types of attacks.
The results exhibit superior accuracy and resilience
against adversarial attacks,
all while maintaining comparatively low computational complexity.
Among these,
on the Non-IID CIFAR-10 dataset,
our algorithm exhibited performance enhancements
of 19.5\% and 14.5\%,
respectively,
in comparison to the state-of-the-art algorithm
when dealing with two types of data poisoning attacks.
\end{abstract}

\section{Introduction}

The constantly increasing amount of data
and the intricacy of machine learning models
have given rise to an augmented requirement
for computational resources.
The current machine learning paradigm
necessitates data collection in a central server,
which may be unfeasible or undesirable
from the perspectives of privacy, security, regulation, or economics.
Recently,
federated learning (FL) \cite{FL}
and other decentralized machine learning methods
\cite{PasquiniRT23,JiangXXWQH23}
have been proposed as potential solutions
to address the above issues.
In a federated learning framework,
the server broadcasts the shared global model to clients,
clients then perform training with their private data sets
and upload the updated models to the server.
Then the server updates the global model
by aggregating the locally updated models
and begins the next round of training.
FedAvg \cite{Fedavg},
which takes the average of local parameters as global parameters,
is a typical aggregation algorithm.

However,
due to its special framework,
the typical FL algorithm faces some serious security threats to the model
if some clients are malicious.
Byzantine failure is one most important security threats,
where some clients are malicious and take measures
to attack the global model.
For example,
malicious clients could upload modified parameters
while genuine clients upload local parameters.
Then, the global model performance
would be degraded even though only one single malicious client
in FedAvg algorithm \cite{attack_effect}.
The untargeted attack \cite{untargeted_attack_1,untargeted_attack_2}
is a model poisoning attacks,
where malicious clients directly manipulate their local updates
to compromise the global model performance.
Shafah~\etal{} \cite{poison_via_feature_collision}
introduced a form of clean-label poisoning attack,
which adds carefully crafted perturbations
to a subset of the training data,
to contaminate an image classification model.
This contamination aims to disrupt the model training process,
leading to training failure.
\cite{
    backdoot_attack,symmetric_attack,pairwise_attack,
    label_flip_7_1_(1),label_flip_7_1_(2)}
conducted investigations
into a backdoor attack strategy named label flipping.
In this attack,
the attacker's model update is engineered
to deliberately induce the local model
to learn an incorrect mapping for a small subset of the data.
As an example,
during training on MNIST,
the attackers may aim to make the model
to classify all images originally labeled ``7'' as ``1'',
thereby obstructing the convergence of the global model.

To address the above issue,
previous research has proposed some robust aggregation rules
to defend against such attacks.
Krum and Multi-Krum \cite{Krum&Multi-Krum},
Median and Trimmed-mean \cite{median&trimmed-mean},
GeoMed \cite{Geomed}, Bulyan \cite{Bulyan}, MAB-RFL \cite{MAB-RFL}
aim to drop potential malicious updates
by comparing local updates.
While \cite{FLTrust,DGDA,Zeno-2019,autoencoder}
distinguish malicious clients
by comparing the local updates and server update,
which is trained using clean data stored on the server.
Hsieh~\etal{} \cite{Non-iid}
discovered that the Non-IID
(not identically and independently distributed) distribution
can notably stymie model training convergence
in various distributed algorithms,
including FedAvg,
yet previous work ignores the effects
of such heterogeneous data distributions.
In the real-world FL system implementation,
we neither know how the attacker is attacking,
nor how many attackers.
And for privacy reasons,
it could be difficult to curate a global dataset
sampled from the underlying data distribution.
An urgent need thus arises
to propose an aggregation rule
that does not require such prior knowledge
and can still effectively defend
against multiple attacks.

In this paper,
we propose a new Robust FL method called \method{}.
The server maintains a credibility set of clients
based on cosine similarity,
which makes the server consider the historical contribution of clients
when assigning weights.
We performed a comprehensive series of experiments
on the MNIST and CIFAR-10 datasets,
aiming to compare the performance of our \method{}
with other existing algorithms.
Our assessment encompassed multiple attack types,
varying fractions of malicious clients and dataset distributions.
The empirical findings unequivocally showcase
that our algorithm not only maintains high test accuracies
but also demonstrates exceptional robustness
against adversarial attacks.
Importantly,
these achievements are coupled with the benefit
of maintaining a comparatively low computational complexity
when contrasted with alternative algorithms.

The main contributions of this work can be summarized as follows:
\begin{itemize}

    \item \textbf{\method{} Scheme:}
    In general,
    this paper proposes a new framework for robust federated learning
    based on credibility values called \method{}
    to address the challenges faced above.
    It maintains and employs a credibility set,
    which weighs the historical clients' contributions
    based on the similarity between the local models and global model,
    to adjust the global model update.
    The elegance of \method{} lies in its inherent ability
    to dynamically adjust the weights of credibility values,
    all while maintaining a commendable computational complexity
    of \( O(n) \).
    Notably,
    this operation is achieved
    without requiring the server
    to possess a training dataset.

    \item \textbf{Credibility Management for FL:}
    In this work,
    we intricately devised a management approach
    for credibility values within \method{}.
    This is achieved through initialization
    and during the training process
    by comparing the similarity
    between the weights of global and local models
    and incorporating a temporal decay function
    to update the reputation values of each client.
    We designed a method that is characterized
    by its low computational complexity,
    while also being capable of efficiently training a global model
    without the necessity of prior knowledge
    regarding the credibility of individual clients.

    \item \textbf{Detailed Evaluation:}
    We conducted extensive comparative experiments
    between \method{} and various algorithms
    proposed in prior research.
    These experiments encompassed diverse datasets,
    disparate distributions, varying attack methods,
    and differing numbers of attackers.
    The results of these experiments demonstrate
    that our algorithm
    attains or surpasses the state-of-the-art algorithms
    in almost all scenarios.
    Notably,
    on the Non-IID CIFAR-10 dataset,
    our algorithm exhibited performance enhancements
    of \( 19.5\% \) and \( 14.5\% \), respectively,
    in comparison to the state-of-the-art algorithm
    when dealing with two types of data poisoning attacks
    as evaluated in our experiments involving four attackers.
    We also tracked the dynamic changes
    in credibility values among different clients
    during the training process.
    The outcomes indicate
    that \method{} effectively distinguishes attackers
    within the client population.

\end{itemize}

The remainder of this paper is structured as follows.
The system model, threat model and defense model
are presented in Section 2,
while Section 3 provides details of our proposed solution.
Experimental results are showcased in Section 4.
The related work and motivation are introduced in Section 5.
Finally, Section 6 concludes the paper.

\section{FL System}

\begin{figure}[htbp]
    \centering
    \includegraphics[width=0.85\textwidth]{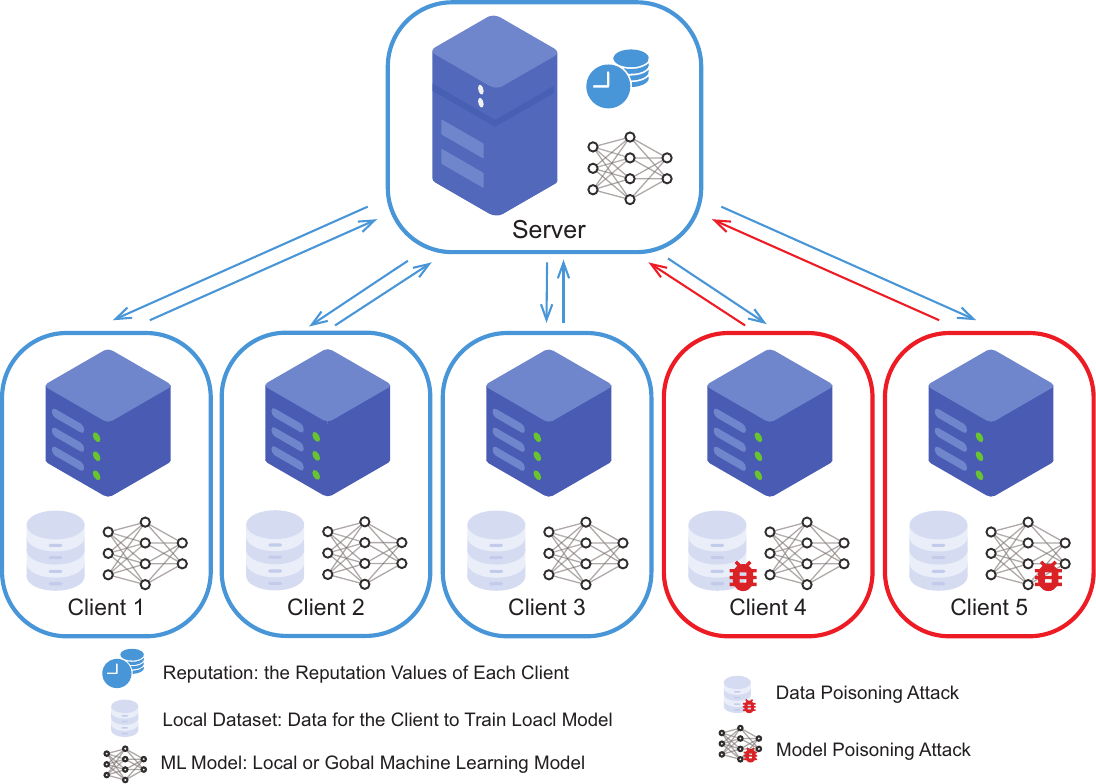}
    \caption{FL system framework}
    \label{fig:FL system framework}
\end{figure}
In this section,
we present the system model, threat model,
and defense model of the Federated Learning (FL) system
considered in this work.
\Cref{fig:FL system framework}
provides an overview of the system.

\subsection{System Model}

We consider a Federated Learning (FL) system
that consists of multiple clients
and one central server
for collaborative model training.
The server forms a global model
by aggregating model parameters
uploaded by clients.
However,
some of these clients
may be compromised and controlled
by malicious attackers,
turning them into adversarial clients \Cref{fig:FL system framework}.
Therefore,
our objective is to find an effective method
to aggregate the model parameters
provided by both the potentially adversarial
and the regular clients,
while maintaining the efficacy
of the global model.
Further details are provided below.

\textbf{Client:}
In an FL system,
the clients refer to the individual participants or devices
that contribute to the collaborative model training.
Each client has its own local dataset and model,
which it uses to train the model.
Before each training session,
clients will receive the latest global model from the server.
They then use their own local datasets
to train respective models
based on this global model.
Finally,
the trained models will be uploaded to the server.

\textbf{Server:}
The server in this system
maintains the credibility values
of each client and the global model.
At the start of each training round,
the server distributes the current global model
to all clients.
Once the clients complete the predetermined number
of training epochs,
the server collects the models trained by each client
and aggregates them into a new global model
using a credibility value-based algorithm.
Subsequently,
the server updates the credibility values of different clients
based on the new global model and the local models
submitted by the clients.
Finally,
the server sends the updated global model
back to the clients
for the subsequent training round.
This process repeats
until reaching the desired accuracy
or the maximum number of training rounds.

\subsection{Threat Model}

In this section,
we present a comprehensive threat model
for poisoning attacks in Federated Learning (FL) system
(\Cref{fig:FL system framework}).
This model encompasses various aspects
such as the objectives of poisoning attacks,
the types of poisoning attacks,
the knowledge possessed by the poisoning attacker,
and the assumptions made for poisoning attacks.

\textbf{Objective of Poisoning Attacker: }
The attacker typically compromises the learning system
to cause failure on specific inputs
intentionally chosen by the attacker.
This process can even construct backdoors
through which they can control the output
of a deployed model in the future.
Similar to numerous prior studies
on poisoning attacks
\cite{fang2020local,sagar2023poisoning,xie2020fall},
we assume that the primary objective
of poisoning attacker in FL system
is to deliberately manipulate the local training process
and compromise the aggregation process of the global model.
The aim is to cause a significant increase
in error rates for the testing data,
thereby undermining the integrity and reliability of the model.

\textbf{Types of Poisoning Attacks: }
Poisoning attacks in FL systems
can be categorized based on their attack methods,
mainly including data poisoning attacks \cite{biggio2012poisoning}
and model poisoning attacks \cite{sagar2023poisoning}.
Data poisoning attacks
involve injecting poisoned data samples into the training dataset,
such as label-flipping attacks \cite{biggio2012poisoning}.
In this paper,
we apply label-flipping attacks
using the pairwise \cite{pairwise_attack}
and symmetric \cite{symmetric_attack} matrix methods.
Model poisoning attacks
aim to target the model parameters directly,
manipulating the aggregation process
by sending error or noisy parameters.
Model poisoning attacks
\cite{fang2020local,like_attacks,An_Experimental_Study}
aim to thwart the FL process
by uploading to the server
either constant model weights,
or weights sampled from a certain distribution,
or parameters that are opposite to the training results.

\textbf{Poisoning Attacker's Knowledge:}
As poisoning attackers
are part of the FL system,
they possess a certain level of knowledge
about the FL system and its components.
This knowledge includes an understanding
of the training data distribution,
the model architecture, the learning algorithm,
and the global model parameters
that are updated during the iterative process
in communication rounds.
However,
conducting data and model poisoning attacks
does not necessarily depend on this knowledge
but rather on the attackers' ability
to collaborate with each other.
For instance,
to execute a model poisoning attack,
several attackers may need
to provide identical constants or distributions
while returning locally trained parameters.

\textbf{Poisoning Attacks Assumptions:}
Poisoning attacks in FL system
are built upon certain assumptions.
\textit{(1)}
Multiple poisoning attackers
can assume the ability to collaborate.
This collaboration
can involve the use of different strategies
by different attackers,
such as pairwise and symmetric matrix label flipping attacks,
or utilize the same model poisoning attacks.
\textit{(2)}
It is assumed that the number of malicious clients
does not exceed half of the total \cite{fang2020local}.
\textit{(3)}
We finally assume that the communication
between the server and client is reliable,
which means that in our paper,
we do not consider noise and errors
caused during the transmission process.

\subsection{Defense Model}

\textbf{Defender's knowledge:}
The defense is performed on the server.
In practical applications,
the server has several limitations.
\textit{(1)}
The server does not possess any training data
on the server side.
\textit{(2)}
The server cannot access the data stored in clients.
\textit{(3)}
The server does not know the number of malicious clients,
and their attack strategy choices.
\textit{(4)}
The server only has access to local model updates.

\vspace{1em}
In this work,
we consider the general FL framework,
which consists of a server and \( n \) clients,
and \( f \) (where \( f < \frac{n}{2} \)) of them
are malicious clients.
The server has \( \rounds \) synchronous rounds
and clients cluster \( C \) has \( E \) local epochs
in each round.
During each round,
the server broadcasts global parameters \( g \)
to the clients.
The clients subsequently train the model
using a mini-batch size of \( B \)
with their respective local datasets
for \( E \) epochs
to obtain local updated model \( g_{i} \).
Among the \( n \) clients,
\( f \) malicious clients perform attacks
by poisoning datasets (data poisoning)
or manipulating local models (model poisoning).
The server collects local updates
and computes a new global model
by applying an aggregation rule.
Detailed definitions of all symbols
are given in \Cref{tab:notations}.
\begin{table}[h]
    \caption{Table of Notations.}%
    \label{tab:notations}
    \begin{tabular}{cl}
        \toprule
        Parameter & Description \\
        \midrule
        \( \alpha_{1},\alpha_{2},\beta \) & Hyperparameters of \method{} \\
        \( \alpha \) & The equilibrium factor \\
        \( n \) & Number of clients \\
        \( f \) & Number of malicious clients (\( 0 \leq f < n / 2 \)) \\
        \( \lr \) & Learning rate \\
        \( B \) & Batch Size \\
        \( C_{i} \) & The \( \ordinal{i} \) client (\( 1 \leq i \leq n \)) \\
        \( \globalmodel \) & The global model \\
        \( \localmodel \) & The local model of \( C_{i} \) \\
        \( \credvals \) & The credibility set of clients \\
        \( \credval \) & The credibility of \( C_{i} \) \\
        \( \credscore \) &
            The credibility score
            between local model of \( C_{i} \) and the global model \\
        \( w_{i} \) & The weight of \( C_{i} \) \\
        \( \rounds \) & Number of global training rounds \\
        \( E \) & Number of local training epochs \\
        \bottomrule
    \end{tabular}
\end{table}
\begin{figure}[htbp]
    \centering
    \includegraphics[width=0.95\textwidth]{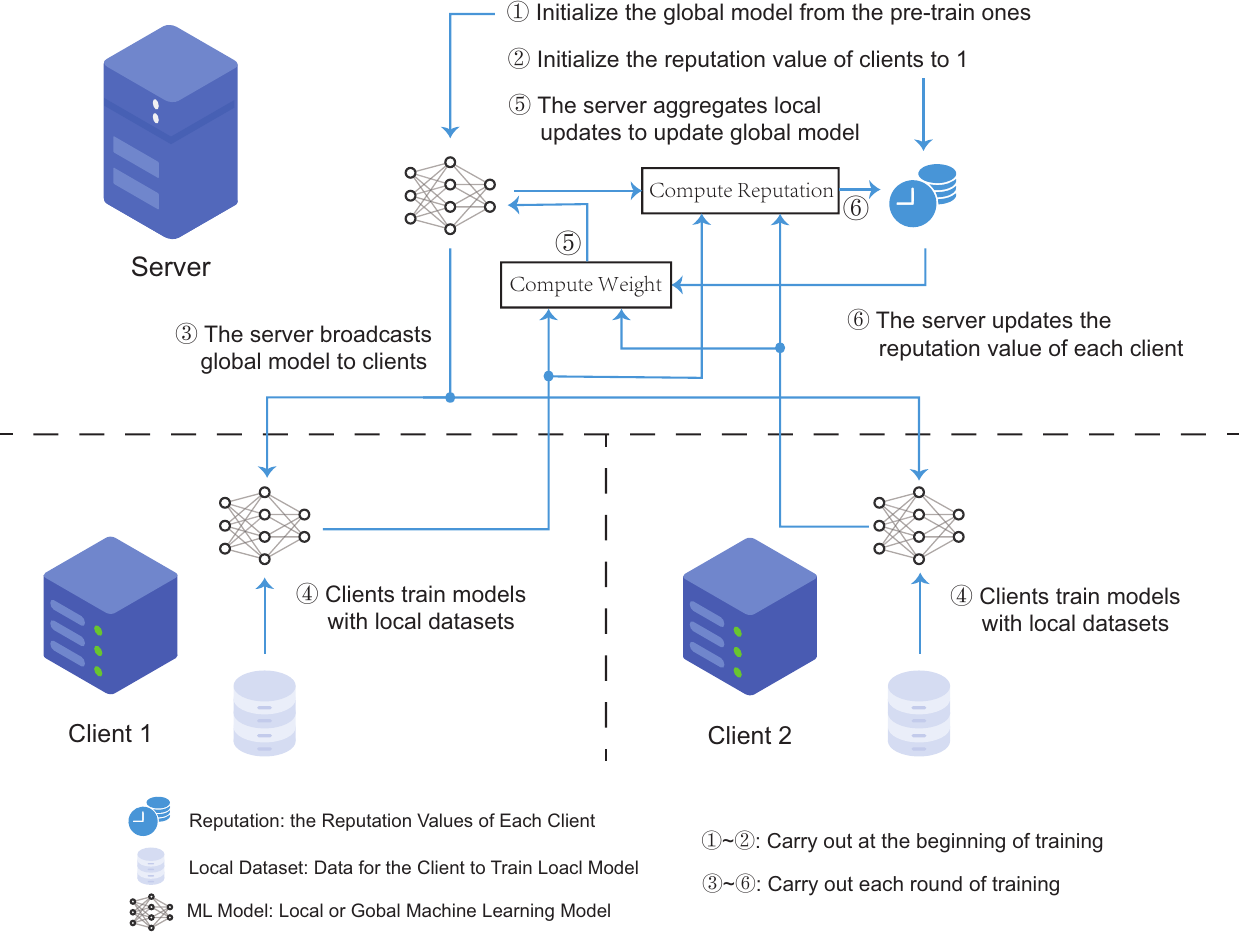}
    \caption{A high-level overview of \method{}.}%
    \label{fig:overview}
\end{figure}

\section{\method{}}

FL algorithms, for instance, FedAvg,
typically aggregates the local model updates from all clients
by averaging the local model updates
to update the global model.
The weight of each client is usually set
to be equal to the size of its local dataset.
In the presence of malicious clients,
the global model is vulnerable to attacks;
and the server faces challenges in accurately determining
whether the clients' data distributions
exhibit natural heterogeneity
or are intentionally manipulated.
This difficulty arises from the server's inability
to sample directly from the clients' datasets.
It is thus preferred
to evaluate the credibility of clients
as each client's weight,
thus reducing the weights of malicious clients
to protect global model training.

To address this issue,
we propose \method{},
a novel defense framework based on credibility management.
It aims to resist various types of attacks
without the knowledge of the number of malicious clients.
The server considers the cosine similarity
of each client's local update
with the global update
and the credibility value of the previous rounds
to update the client's credibility.

Our approach envisions the server
dynamically managing a set of credibility values for each client.
These values assess historical contributions
by considering the similarity
between local models and the global model.
Incorporating temporal decay and credibility values,
the server judiciously adjusts the weights
assigned to each client's local updates
for global model aggregation.
The overall algorithm
of our proposed \method{} method
can be found in \Cref{alg:Fed-credit},
and it is summarized in the following five steps:
\begin{itemize}

    \item \textbf{Step 1:}
    The server initializes the global model \( \globalmodel \)
    and assigns an initial credibility value of 1 to each client.
    Next, the server iterates Steps 2 to 5
    until either the global model \( \globalmodel \)
    achieves the desired performance
    or reaches the maximum allowable number
    of global training epochs \( \rounds \).

    \item \textbf{Step 2:}
    The server sends the global model \( \globalmodel \)
    to all clients.

    \item \textbf{Step 3:}
    The clients \( C_{i} \) independently train models
    using their own local datasets.

    There is no communication between benign clients,
    ensuring that they cannot exchange the datasets
    or trained models with other clients.
    At the end of the training process,
    the clients upload the model parameters \( \localmodel \)
    to the server.

    \item \textbf{Step 4:}
    The server incorporates an equilibrium factor \( \alpha \)
    to dynamically regulate the impact
    of the credibility value \( \credval \)
    on the client weights \( w_{i} \).
    \( \alpha \) is obtained in \Cref{alg:Weight}.
    As training progresses,
    the influence of the credibility value \( \credval \)
    on weights \( w_{i} \) gradually increases.
    The server then aggregates the local updates
    using the client weights \( w_{i} \). 

    \item \textbf{Step 5:}
    The server assesses the cosine similarity
    between individual layers
    of each local model \( \localmodel \)
    and the global models
    to obtain the credibility score \( \credscore \)
    of \( localmodel \).
    \method{} takes 
    the decaying effect over time
    into consideration,
    thus an exponential decay factor 
    is utilized
    to average historical credibility values.
    Finally,
    the server normalizes the credibility values
    of each client.
\end{itemize}

\begin{algorithm}
\caption{The \method{} algorithm.}\label{alg:Fed-credit}
\begin{algorithmic}[1]
\Require{%
    Clients with local training datasets,
    \( \braces{C_{1},C_{2},C_{3},\ldots,C_{n}} \);
    learning rate \( \lr \);
    batch size \( B \);
    number of local training iterations \( E \);
    number of communications \( \rounds \);
    hyperparameters \( \alpha_{1}, \alpha_{2}, \beta \).}
\Ensure{Global model \(\globalmodel\).}
\State{%
    // \textbf{Step 1:}
    The server initiates the global model and the credibility values.
}
\State{Initialize \( \globalmodel \)}
\State{\( \credvals \gets \mathbf{1} \).}
\Comment{%
    \(
        \credvals = \bracks{
            \tau_{1},\tau_{2},\tau_{3},\ldots,\tau_{n}
        }
    \)
    contains the credibility value of clients.
}
\For{\( r \in \rounds \)}
    \State{%
        // \textbf{Step 2:}
        The server broadcasts global model \( \globalmodel \).
    }
    \State{%
        The server sends global model \( \bm{g } \)
        to all clients \( \braces{C_{1},C_{2},C_{3},\ldots,C_{n}} \).
    }
    \State{%
        // \textbf{Step 3:}
        Clients train models with local datasets.
    }
    \For{\( i = 1 \) to \( n \)}
        \Comment{do in parallel}
        \State{\(
            \bm{g_{i} } \gets
                \textbf{getLocalModel}\parens{
                    \globalmodel, C_{i}, \lr, E, B
                }
        \)}
        \State{Return \( \bm{g_{i} } \) to server.}
    \EndFor
    \State{%
        // \textbf{Step 4:}
        The server aggregates local updates
        to update the global model.
    }
    \For{\( i = 1 \) to \( n \)}
        \State{\(
            w_{i} \gets \textbf{getWeight}\parens{
                \rounds, \credval, \alpha_{1}, \alpha_{2}
            }
        \)}
        \Comment{%
            \( \credval \) is the credibility value
            of the \( \ordinal{i} \) client
        }
    \EndFor
    \State{\(
        \globalmodel \gets
            \sum_{i=1}^{n} w_{i} \cdot \localmodel
    \)}
    \Comment{Combine local gradients}
    \State{// \textbf{Step 5:}}
        The server updates the credibility value of each client.
    \For{\( i = 1 \) to \( n \)}
        \Comment{do in parallel}
        \State{\(
            \credval \gets \textbf{getCredibility}\parens{
                \credval, \localmodel, \globalmodel, \beta
            }
        \)}
    \EndFor
    \State{\(
        \credvals \gets \credvals - min(\credvals)
    \)}
\EndFor
\State{%
    \textbf{Return} the global model \( \globalmodel \)
}
\end{algorithmic}
\end{algorithm}

\subsection{Credibility Management Mechanism}

In a FL system,
users with higher credibility
and more stable network connections
contribute more to the training process.
On the FL system initialization,
it assigns each client
a credibility value \( \credval = 1 \).
After each round of training,
it aggregates the client model parameters \( \localmodel \)
into the global model,
where the aggregation weight \( w_i \) of each client
is determined by its credibility value \( \credval \).
We then evaluate the credibility score \( \credscore \),
which is computed
by averaging the cosine similarity of each layer
between local and global models.
The algorithm for credibility value assignment
is outlined in \Cref{alg:credibility}.

The credibility value \( \credval \)
is then updated with the credibility score \( \credscore \)
using an exponential decay function
to take
the decaying effect over time
into consideration.
Thus
an decay factor \( \beta \)
is utilized 
to average 
historical credibility values.
A larger \( \beta \) value
signifies that the past credibility value
holds lesser significance,
thereby highlighting the augmented significance
of the current credibility score \( \credscore \).
Following this,
to mitigate the attacks
from malicious clients,
we normalize credibility values
by subtracting the minimum credibility value
from all values:
\begin{equation}
    \credvals = \credvals - \min\parens{\credvals}
    \label{eq:cred_norm}
\end{equation}

\begin{algorithm}
\caption{%
    The getCredibility function.
}\label{alg:credibility}
\begin{algorithmic}[1]
\renewcommand{\thefootnote}{\ding{\numexpr192+\value{footnote}}}
\Require{%
    Credibility value \( \credval \);
    local model \( \localmodel \);
    global model \( \globalmodel \),
    hyperparameters \( \beta \)}
\Ensure{Updated credibility value \( \credval \)}
\State{%
    // Compute the credibility score \( \credscore. \)}
\State{\( \credscore \gets 0 \)}
\Comment{Initialize the credibility score}
\For{\( layer \)\footnotemark[1] in \( \globalmodel \)}
    \State{\(
        \credscore +=
            {
                \angles{
                    \localmodel[layer],
                    \globalmodel[layer]
                }
            } / {
                \norm{\localmodel[layer]}
                \norm{\globalmodel[layer]}
            }
    \)}
    \Comment{The sum of cosine similarity of each layer}
\EndFor
\State{%
    // Utilize the \( \credscore \)
    and historical credibility value
    to update the new credibility value \( \credval \).
}
\State{\(
    \credval =
        \beta \cdot \credscore +
        (1 - \beta) \cdot \credval
\)}
\State{\textbf{Return} \( \credval \)}
\end{algorithmic}
\end{algorithm}
Note that in \Cref{alg:credibility},
the bias parameters in each layer are also used
to evaluate the \( \credscore \) of the model.

\subsection{Updating Weight}
Initially,
the preference of the proposed scheme
is to assign uniform weights \( w_{i} \)
to all clients
in order to prevent inadvertent misclassification
of malicious clients.
Subsequent to this initial phase,
our aim is to identify malicious clients
through their credibility values \( \credval \)
and significantly reduce their weights \( w_{i} \)
during training
to protect the global model \( \globalmodel \).
In pursuit of this objective,
we introduced the equilibrium factor \( \alpha \),
which is calculated
based on a variant sigmoid function \eqref{eq:sigmoid}.
In our weight formula \eqref{eq:weight},
we utilized it
to dynamically modulate the significance
of both the credibility value \( \credval \)
and the average value
within the weights \( w_{i} \).
The comprehensive weight calculation procedure
is illustrated in \Cref{alg:Weight}.
\begin{equation}
    \alpha = \parens*{
        1+e^{\parens*{
            -(\rounds+\alpha_{1}) / \alpha_{2}
        }}
    }^{-1}
    \label{eq:sigmoid}
\end{equation}
\begin{equation}
    w_{i} =
        \frac{1}{n} \cdot (1 - \alpha) +
        \frac{\credval}{\sum_{j = 1}^{n}\tau_{j}} \cdot \alpha
    \label{eq:weight}
\end{equation}
\begin{algorithm}
\caption{getWeight} \label{alg:Weight}
\begin{algorithmic}[1]
    \Require Credibility value \( \credval \);
        number of training rounds \( \rounds \);
        hyperparameters \( \alpha_{1},\alpha_{2} \).
    \Ensure The weight of \( \ordinal{i} \) client \( w_{i} \).
    \State{
        //
        Compute the equilibrium factor \( \alpha \)
        by the variant sigmoid function.}
    \State{\(
        \alpha = \parens*{
            1+e^{\parens*{
                -(\rounds+\alpha_{1}) / \alpha_{2}
            }}
        }^{-1}
    \)}
    \State{// Update the weight \( w_{i} \)}
    \State{\(
        w_{i} =
            \frac{1}{n} \cdot (1 - \alpha) +
            \frac{\credval}{
                \sum_{j = 1}^{n}\tau_{j}
            } \cdot \alpha
    \)}
    \State{\textbf{Return} \( w_{i} \)}
\end{algorithmic}
\end{algorithm}

\section{Experiment Result}

\subsection{Experiment Setup}

Our experimental platform comprises the AMD EPYC 7742 64-Core Processor and the NVIDIA Tesla A100 40G computing accelerator. We conducted a comparative analysis of our approach, \method{}, with several existing methods including FedAvg \cite{Fedavg}, GeoMed \cite{Geomed}, Krum \cite{Krum&Multi-Krum}, Median \cite{median&trimmed-mean}, Multi-Krum \cite{Krum&Multi-Krum}, Trimmed \cite{median&trimmed-mean}, and FLTrust \cite{FLTrust}. This evaluation was carried out on the MNIST and CIFAR-10 datasets, considering varying numbers of attackers as well as both iid and Non-iid data distribution settings. For MNIST, we choose a Multi-Layer Perceptron (MLP) network with two hidden layers and one output layer to train the global model. For CIFAR-10, We opt for a lightweight model called Compact Convolutional Transformers (CCT) \cite{CCT}, as its small size and effectiveness offer better potential in addressing the resource constraints of onboard FL devices. We utilized Dirichlet distribution to model Non-iid distribution \cite{fed-lab}. The hyperparameter settings of this work are shown in \Cref{hyperparameters}.

\textbf{MNIST:} The MNIST dataset is a well-known collection of handwritten digits widely used in the field of machine learning. It consists of 60,000 training examples and 10,000 testing examples. Each image is a 28x28 grayscale image of a digit, ranging from 0 to 9. The MNIST dataset serves as a benchmark for evaluating image classification algorithms and has played a crucial role in advancing the field of deep learning.

\textbf{CIFAR-10:} The CIFAR-10 dataset is a popular benchmark dataset in the field of computer vision. It consists of 60,000 color images, each of size 32x32 pixels, divided into 10 different classes. The dataset serves as a standard evaluation tool for image classification algorithms and has played a significant role in advancing the field of deep learning.

\textbf{Attack types:} In our experiments, we mainly use two attack methods, data poisoning attacks and model poisoning attacks. Among the data poisoning attacks, we select the label flipping attack based on pairwise (PW) and symmetric (SM) matrices. As for model poisoning attacks, we have chosen three different implementations. Specifically, Constant Parameter (CP), where all model parameters are identical; Normal Parameter (NP), where returned model parameters follow a normal distribution; and Sign-Flip Parameter (SF), which returns a model with parameters opposite to those obtained during training.

\textbf{Evaluation:} To evaluate the multiple defense model, as many other works \cite{Manipulating} \cite{An_Experimental_Study} \cite{Challenges_and_Approaches}, we adopted the accuracy as a criterion. The accuracy is employed to judge which represents the proportion of correctly classified samples to the total number of samples in the test dataset and is defined in Eq. (\ref{accuracy}).

\begin{equation}
    \text{accuracy} = \frac{TP + TN}{TP + FN + FP + FN}
    \label{accuracy}
\end{equation}

\begin{table}[]
\begin{tabular}{||c|c|cc||}
\hline
Parameter            & Description                                      & \multicolumn{2}{c||}{Value}                                                            \\ \hline \hline
\( n \)                  & Number of clients                                & \multicolumn{2}{c||}{10}                                                               \\ \hline
\multirow{2}{*}{\( f \)} & \multirow{2}{*}{Number of malicious clients}     & \multicolumn{1}{c|}{\begin{tabular}[c]{@{}c@{}}Model\\ Poison\end{tabular}} & 1, 2, 3 \\ \cline{3-4}
                     &                                                  & \multicolumn{1}{c|}{\begin{tabular}[c]{@{}c@{}}Data\\ Poison\end{tabular}}  & 1, 2, 4 \\ \hline
\( \lr \)                 & Learning rate                                    & \multicolumn{2}{c||}{0.01}                                                             \\ \hline
\multirow{2}{*}{\( B \)} & \multirow{2}{*}{Batch size}                      & \multicolumn{1}{c|}{MNIST}                                                  & 64      \\ \cline{3-4}
                     &                                                  & \multicolumn{1}{c|}{CIFAR-10}                                               & 32      \\ \hline
\( \rounds \)             & Number of global training epochs                 & \multicolumn{2}{c||}{100}                                                              \\ \hline
\multirow{2}{*}{\( E \)} & \multirow{2}{*}{Number of local training epochs} & \multicolumn{1}{c|}{MNIST}                                                  & 5       \\ \cline{3-4}
                     &                                                  & \multicolumn{1}{c|}{CIFAR-10}                                               & 2       \\ \hline
\( \alpha_{1} \)         & \multirow{3}{*}{Hyperparameters of \method{}}     & \multicolumn{2}{c||}{1}                                                                \\ \cline{1-1} \cline{3-4}
\( \alpha_{2} \)         &                                                  & \multicolumn{2}{c||}{0.8}                                                              \\ \cline{1-1} \cline{3-4}
\( \beta \)              &                                                  & \multicolumn{2}{c||}{0.1}                                                              \\ \hline
\end{tabular}
\vspace{0.5em}
\caption{Hyperparameters settings}
\label{hyperparameters}
\end{table}

\subsection{Numerical Analysis}
\begin{table}[]
\resizebox{0.915\textwidth}{!}{
\begin{tabular}{|c||c||c||c||c||c|c|c|c|c|c|c|c|}
\hline
Dataset & Distribution & \begin{tabular}[c]{@{}c@{}}Attack\\ type\end{tabular} &  & f & Fed-credit & FedAvg & GeoMed & Krum & Multi-Krum & Median & Trimmed & FLTrust \\ \hline \hline
 &  & No attack & - & 0 & \textbf{97.61} & 97.63 & \textbf{97.81} & 95.64 & 96.33 & 97.64 & 97.59 & 86.85 \\ \cline{3-13}
 &  &  &  & 1 & \textbf{97.55} & 90.18 & 97.62 & 95.53 & 96.37 & \textbf{97.66} & 97.49 & 86.53 \\ \cline{5-13}
 &  &  &  & 2 & \textbf{97.36} & 83.09 & 97.34 & 95.56 & 96.32 & \textbf{97.50} & 88.34 & 86.58 \\ \cline{5-13}
 &  &  & \multirow{-3}{*}{\begin{tabular}[c]{@{}c@{}}CP\end{tabular}} & 3 & \textbf{97.07} & 67.38 & \textbf{97.17} & 95.17 & 96.29 & 97.04 & 71.97 & 85.79 \\ \cline{4-13}
 &  &  &  & 1 & \textbf{97.55} & 91.82 & 97.61 & 94.87 & 96.33 & \textbf{97.63} & 97.61 & 85.76 \\ \cline{5-13}
 &  &  &  & 2 & \textbf{97.42} & 88.39 & 97.39 & 94.93 & 96.43 & \textbf{97.49} & 90.55 & 86.37 \\ \cline{5-13}
 &  &  & \multirow{-3}{*}{\begin{tabular}[c]{@{}c@{}}NP\end{tabular}} & 3 & \textbf{97.22} & 84.48 & \textbf{97.31} & 95.20 & 96.44 & 97.27 & 85.34 & 86.96 \\ \cline{4-13}
 &  &  &  & 1 & \textbf{97.55} & 88.18 & 97.48 & 95.37 & 96.49 & 97.49 & \textbf{97.55} & 86.78 \\ \cline{5-13}
 &  &  &  & 2 & \textbf{97.28} & 76.42 & 97.47 & 94.53 & 96.45 & \textbf{97.53} & 85.27 & 85.22 \\ \cline{5-13}
 &  & \multirow{-9}{*}{\begin{tabular}[c]{@{}c@{}}Model\\ poison\end{tabular}} & \multirow{-3}{*}{\begin{tabular}[c]{@{}c@{}}SF\end{tabular}} & 3 & \textbf{97.23} & {\color[HTML]{9B9B9B} 11.35} & 97.19 & 95.48 & 96.36 & \textbf{97.44} & 72.01 & 87.67 \\ \cline{3-13}
 &  &  &  & 1 & \textbf{97.47} & 97.51 & 97.59 & 94.88 & 96.39 & \textbf{97.59} & 97.52 & 87.64 \\ \cline{5-13}
 &  &  &  & 2 & \textbf{97.47} & 96.45 & \textbf{97.48} & 94.73 & 96.31 & 97.38 & 97.19 & 85.98 \\ \cline{5-13}
 &  &  & \multirow{-3}{*}{PW} & 4 & \textbf{97.17} & 81.22 & 96.81 & 95.22 & 96.46 & 96.47 & 85.33 & 85.54 \\ \cline{4-13}
 &  &  &  & 1 & \textbf{97.60} & 97.56 & \textbf{97.80} & 94.98 & 96.33 & 97.58 & 97.62 & 86.12 \\ \cline{5-13}
 &  &  &  & 2 & \textbf{97.46} & 97.11 & \textbf{97.51} & 94.83 & 96.50 & 97.56 & 97.14 & 86.02 \\ \cline{5-13}
 & \multirow{-16}{*}{iid} & \multirow{-6}{*}{\begin{tabular}[c]{@{}c@{}}Data\\ poison\end{tabular}} & \multirow{-3}{*}{SM} & 4 & \textbf{97.17} & 93.33 & \textbf{97.25} & 95.19 & 96.34 & 96.95 & 94.65 & 86.19 \\ \cline{2-13}
 &  & No attack & - & 0 & \textbf{96.18} & \textbf{96.43} & 96.23 & 85.30 & 93.08 & 95.52 & 96.12 & 72.26 \\ \cline{3-13}
 &  &  &  & 1 & \textbf{95.91} & 70.87 & 95.49 & 69.13 & 93.24 & 94.71 & 94.98 & 70.45 \\ \cline{5-13}
 &  &  &  & 2 & \textbf{95.76} & 61.95 & 94.79 & 81.06 & 93.23 & 95.29 & 66.76 & 68.84 \\ \cline{5-13}
 &  &  & \multirow{-3}{*}{\begin{tabular}[c]{@{}c@{}}CP\end{tabular}} & 3 & \textbf{95.19} & {\color[HTML]{9B9B9B} 27.31} & 92.71 & 83.88 & 93.16 & 94.15 & 38.47 & 61.70 \\ \cline{4-13}
 &  &  &  & 1 & \textbf{95.71} & 78.47 & 95.53 & 74.57 & 92.89 & 94.41 & 95.27 & 67.44 \\ \cline{5-13}
 &  &  &  & 2 & \textbf{96.29} & 70.93 & 95.49 & 81.07 & 92.16 & 95.24 & 70.84 & 64.10 \\ \cline{5-13}
 &  &  & \multirow{-3}{*}{\begin{tabular}[c]{@{}c@{}}NP\end{tabular}} & 3 & \textbf{95.84} & 57.87 & 93.57 & 81.51 & 92.94 & 95.00 & 54.09 & 67.38 \\ \cline{4-13}
 &  &  &  & 1 & \textbf{95.75} & 67.26 & \textbf{95.85} & 76.50 & 93.18 & 95.02 & 95.26 & 67.86 \\ \cline{5-13}
 &  &  &  & 2 & \textbf{95.90} & 52.29 & 95.57 & 81.41 & 93.26 & 95.84 & 64.41 & 71.58 \\ \cline{5-13}
 &  & \multirow{-9}{*}{\begin{tabular}[c]{@{}c@{}}Model\\ poison\end{tabular}} & \multirow{-3}{*}{\begin{tabular}[c]{@{}c@{}}SF\end{tabular}} & 3 & \textbf{95.48} & {\color[HTML]{9B9B9B} 21.08} & 93.61 & 82.65 & 92.95 & 94.78 & {\color[HTML]{9B9B9B} 27.18} & 70.09 \\ \cline{3-13}
 &  &  &  & 1 & \textbf{95.97} & 95.77 & 96.10 & 72.99 & 93.24 & 94.66 & \textbf{96.21} & 63.02 \\ \cline{5-13}
 &  &  &  & 2 & \textbf{95.89} & 95.60 & 96.33 & 73.79 & 91.90 & 95.23 & 95.87 & 70.08 \\ \cline{5-13}
 &  &  & \multirow{-3}{*}{PW} & 4 & \textbf{94.96} & 74.56 & 94.43 & 75.62 & 93.22 & 93.14 & 79.50 & 59.46 \\ \cline{4-13}
 &  &  &  & 1 & \textbf{96.02} & 95.53 & \textbf{96.17} & 71.73 & 92.88 & 93.76 & 95.42 & 62.27 \\ \cline{5-13}
 &  &  &  & 2 & \textbf{96.23} & 95.60 & \textbf{96.30} & 81.09 & 92.88 & 95.12 & 96.14 & 68.88 \\ \cline{5-13}
\multirow{-32}{*}{MNIST} & \multirow{-16}{*}{Non-iid} & \multirow{-6}{*}{\begin{tabular}[c]{@{}c@{}}Data\\ poison\end{tabular}} & \multirow{-3}{*}{SM} & 4 & \textbf{93.78} & 92.04 & \textbf{95.14} & 70.16 & 85.48 & 93.53 & 93.83 & 52.79 \\ \hline \hline
 &  & No attack & - & 0 & \textbf{68.34} & 68.99 & \textbf{69.41} & 58.20 & 61.09 & 68.08 & 68.72 & 46.91 \\ \cline{3-13}
 &  &  &  & 1 & \textbf{69.55} & 44.68 & \textbf{69.72} & 56.68 & 60.57 & 68.50 & 69.13 & 46.74 \\ \cline{5-13}
 &  &  &  & 2 & \textbf{68.16} & {\color[HTML]{9B9B9B} 15.04} & 68.02 & 57.72 & 61.62 & 68.25 & {\color[HTML]{9B9B9B} 18.50} & 46.77 \\ \cline{5-13}
 &  &  & \multirow{-3}{*}{\begin{tabular}[c]{@{}c@{}}CP\end{tabular}} & 3 & \textbf{65.97} & {\color[HTML]{9B9B9B} 13.78} & 61.97 & 55.79 & 61.82 & 66.43 & {\color[HTML]{9B9B9B} 13.90} & 47.59 \\ \cline{4-13}
 &  &  &  & 1 & \textbf{69.24} & 58.62 & \textbf{68.95} & 56.45 & 62.87 & 67.52 & 69.07 & 46.32 \\ \cline{5-13}
 &  &  &  & 2 & \textbf{68.62} & 37.45 & \textbf{69.02} & 58.17 & 62.90 & 66.70 & 54.96 & 46.87 \\ \cline{5-13}
 &  &  & \multirow{-3}{*}{\begin{tabular}[c]{@{}c@{}}NP\end{tabular}} & 3 & \textbf{68.60} & {\color[HTML]{9B9B9B} 10.00} & 67.93 & 57.11 & 62.12 & 66.02 & {\color[HTML]{9B9B9B} 24.88} & 46.04 \\ \cline{4-13}
 &  &  &  & 1 & \textbf{69.67} & 37.94 & 68.99 & 57.62 & 62.27 & 67.16 & 69.34 & 46.97 \\ \cline{5-13}
 &  &  &  & 2 & \textbf{69.13} & {\color[HTML]{9B9B9B} 10.00} & 68.24 & 58.49 & 62.22 & 65.59 & {\color[HTML]{9B9B9B} 24.65} & 46.24 \\ \cline{5-13}
 &  & \multirow{-9}{*}{\begin{tabular}[c]{@{}c@{}}Model\\ poison\end{tabular}} & \multirow{-3}{*}{\begin{tabular}[c]{@{}c@{}}SF\end{tabular}} & 3 & \textbf{69.40} & {\color[HTML]{9B9B9B} 10.00} & 67.42 & 57.67 & 61.50 & 65.39 & {\color[HTML]{9B9B9B} 10.00} & 46.91 \\ \cline{3-13}
 &  &  &  & 1 & \textbf{68.69} & 67.13 & 68.21 & 54.24 & 62.26 & 67.08 & 68.25 & 45.99 \\ \cline{5-13}
 &  &  &  & 2 & \textbf{68.01} & 64.50 & 66.81 & 57.74 & 62.93 & 64.47 & 65.21 & 47.14 \\ \cline{5-13}
 &  &  & \multirow{-3}{*}{PW} & 4 & \textbf{66.10} & 48.09 & 50.43 & 55.85 & 61.07 & 48.95 & 48.06 & 45.28 \\ \cline{4-13}
 &  &  &  & 1 & \textbf{68.63} & 67.85 & 68.01 & 56.40 & 62.73 & 66.00 & 68.27 & 46.29 \\ \cline{5-13}
 &  &  &  & 2 & \textbf{68.57} & 64.36 & 67.07 & 56.92 & 62.32 & 61.28 & 65.87 & 45.87 \\ \cline{5-13}
 & \multirow{-16}{*}{iid} & \multirow{-6}{*}{\begin{tabular}[c]{@{}c@{}}Data\\ poison\end{tabular}} & \multirow{-3}{*}{SM} & 4 & \textbf{66.70} & 56.98 & 52.39 & 55.41 & 61.96 & 51.28 & 55.92 & 45.82 \\ \cline{2-13}
 &  &  No attack & - & 0 & \textbf{67.68} & 68.64 & \textbf{69.14} & 59.09 & 62.45 & 67.84 & 68.87 & 46.84 \\ \cline{3-13}
 &  &  &  & 1 & \textbf{69.07} & 44.63 & 68.41 & 56.52 & 61.87 & 68.57 & 69.04 & 47.52 \\ \cline{5-13}
 &  &  &  & 2 & \textbf{68.57} & {\color[HTML]{9B9B9B} 15.22} & 67.84 & 57.05 & 60.97 & \textbf{68.68} & {\color[HTML]{9B9B9B} 17.42} & 46.58 \\ \cline{5-13}
 &  &  & \multirow{-3}{*}{\begin{tabular}[c]{@{}c@{}}CP\end{tabular}} & 3 & \textbf{66.42} & {\color[HTML]{9B9B9B} 14.36} & 60.55 & 58.98 & 62.01 & \textbf{67.37} & {\color[HTML]{9B9B9B} 14.08} & 46.56 \\ \cline{4-13}
 &  &  &  & 1 & \textbf{68.86} & 59.27 & \textbf{69.30} & 59.53 & 61.73 & 67.15 & 69.23 & 47.08 \\ \cline{5-13}
 &  &  &  & 2 & \textbf{69.11} & 37.71 & 68.49 & 54.37 & 62.52 & 67.18 & 55.22 & 47.77 \\ \cline{5-13}
 &  &  & \multirow{-3}{*}{\begin{tabular}[c]{@{}c@{}}NP\end{tabular}} & 3 & \textbf{68.65} & {\color[HTML]{9B9B9B} 10.00} & 67.08 & 57.39 & 62.56 & 65.98 & {\color[HTML]{9B9B9B} 25.47} & 45.83 \\ \cline{4-13}
 &  &  &  & 1 & \textbf{69.93} & 37.99 & 69.17 & 58.63 & 61.55 & 67.55 & 68.94 & 46.58 \\ \cline{5-13}
 &  &  &  & 2 & \textbf{69.38} & {\color[HTML]{9B9B9B} 10.00} & 68.94 & 48.99 & 61.84 & 67.43 & {\color[HTML]{9B9B9B} 25.64} & 47.54 \\ \cline{5-13}
 &  & \multirow{-9}{*}{\begin{tabular}[c]{@{}c@{}}Model\\ poison\end{tabular}} & \multirow{-3}{*}{\begin{tabular}[c]{@{}c@{}}SF\end{tabular}} & 3 & \textbf{69.57} & {\color[HTML]{9B9B9B} 10.00} & 67.51 & 55.73 & 60.69 & 64.42 & {\color[HTML]{9B9B9B} 10.00} & 46.34 \\ \cline{3-13}
 &  &  &  & 1 & \textbf{68.32} & 67.33 & 68.13 & 56.98 & 61.39 & 67.31 & 68.27 & 47.11 \\ \cline{5-13}
 &  &  &  & 2 & \textbf{67.82} & 64.31 & 65.89 & 56.61 & 60.75 & 65.90 & 65.94 & 47.20 \\ \cline{5-13}
 &  &  & \multirow{-3}{*}{PW} & 4 & \textbf{66.75} & 47.88 & 49.91 & 54.43 & 62.26 & 49.45 & 48.38 & 48.57 \\ \cline{4-13}
 &  &  &  & 1 & \textbf{68.79} & 67.92 & 68.35 & 54.76 & 61.11 & 66.11 & 68.47 & 46.97 \\ \cline{5-13}
 &  &  &  & 2 & \textbf{67.95} & 66.14 & 66.46 & 45.83 & 61.65 & 62.59 & 65.90 & 46.13 \\ \cline{5-13}
\multirow{-32}{*}{CIFAR-10} & \multirow{-16}{*}{Non-iid} & \multirow{-6}{*}{\begin{tabular}[c]{@{}c@{}}Data\\ poison\end{tabular}} & \multirow{-3}{*}{SM} & 4 & \textbf{66.49} & 55.10 & 52.59 & 55.21 & 62.30 & 52.10 & 55.88 & 47.48 \\ \hline
\end{tabular}
}
\caption{Experiment results overview\\ \small{Constant Parameter (CP), Normal Parameter (NP), Sign Flipping (SF), Pairwise (PW), Symmetric (SM)}}
\label{result}
\end{table}

\Cref{result} shows the accuracy of different robust aggregation rules under various attacks. We can find that \method{} performs well in many situations.

\subsubsection{Impact of Number of Malicious Clients}\

\begin{figure}[htbp]
    \centering
    \includegraphics[width=1\textwidth]{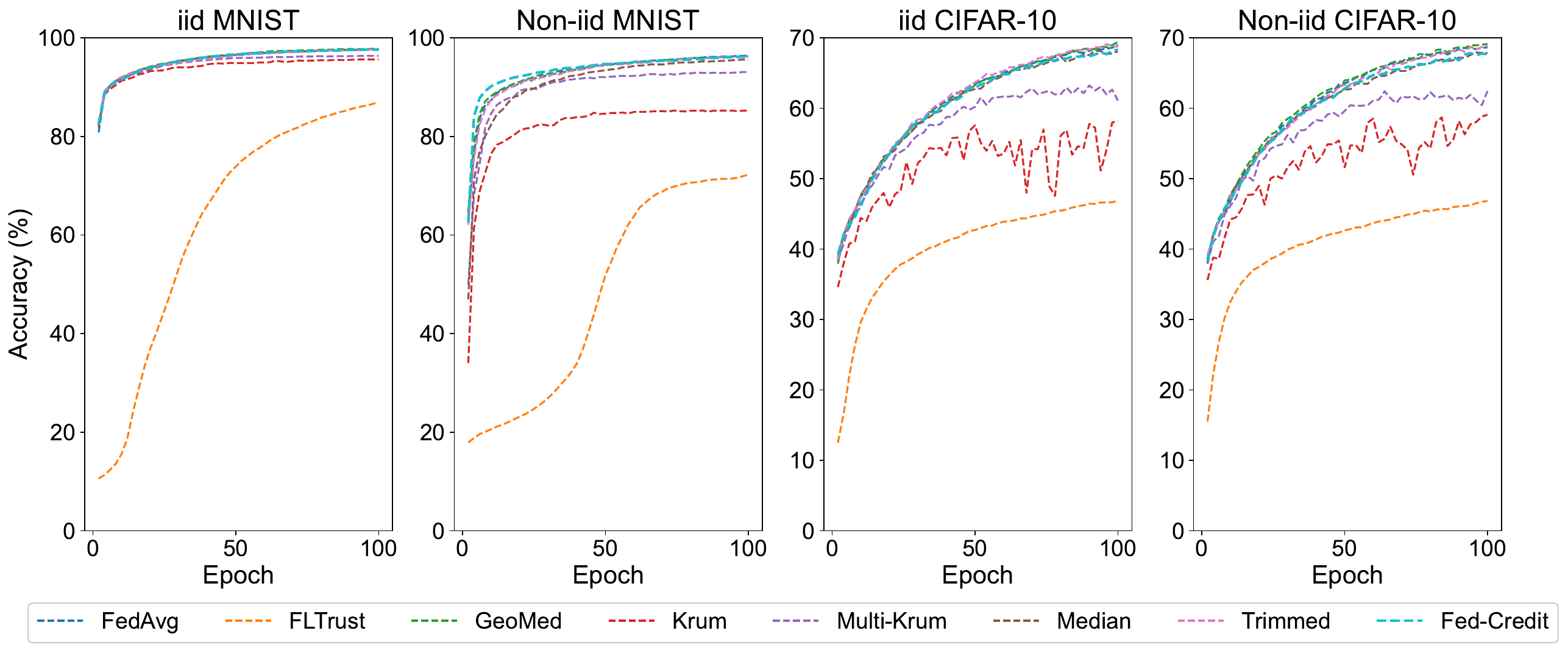}
    \caption{Accuracy without attacks of iid and Non-iid datasets}
    \label{fig:acc-no-attacks}
\end{figure}

First of all, our results demonstrated that with the absence of attacks, \method{}, GeoMed, Median, Trimmed and FedAvg achieve relatively higher accuracy while FLTrust, Krum and Multi-Krum get lower accuracy. Especially, the disparity is more obvious when the dataset distribution is Non-iid. For instance, from the results shown in \Cref{fig:acc-no-attacks}, for MNIST with Non-iid distribution, the accuracy of lower three aggregation rules (FLTrust, Krum, Multi-Krum) are \textbf{72.26\%, 85.30\%, 93.08\%} which are significantly lower than other methods that are around \textbf{96\%}. This might be because Krum and Multi-Krum tend to use one or few local updates to update the global model, which makes the global model cannot fit the overall dataset well. Another finding is that the FLTrust converges slower than other methods, which is consistent with Cao~\etal{} \cite{FLTrust}.

\begin{figure}[htbp]
    \centering
    \includegraphics[width=1\textwidth]{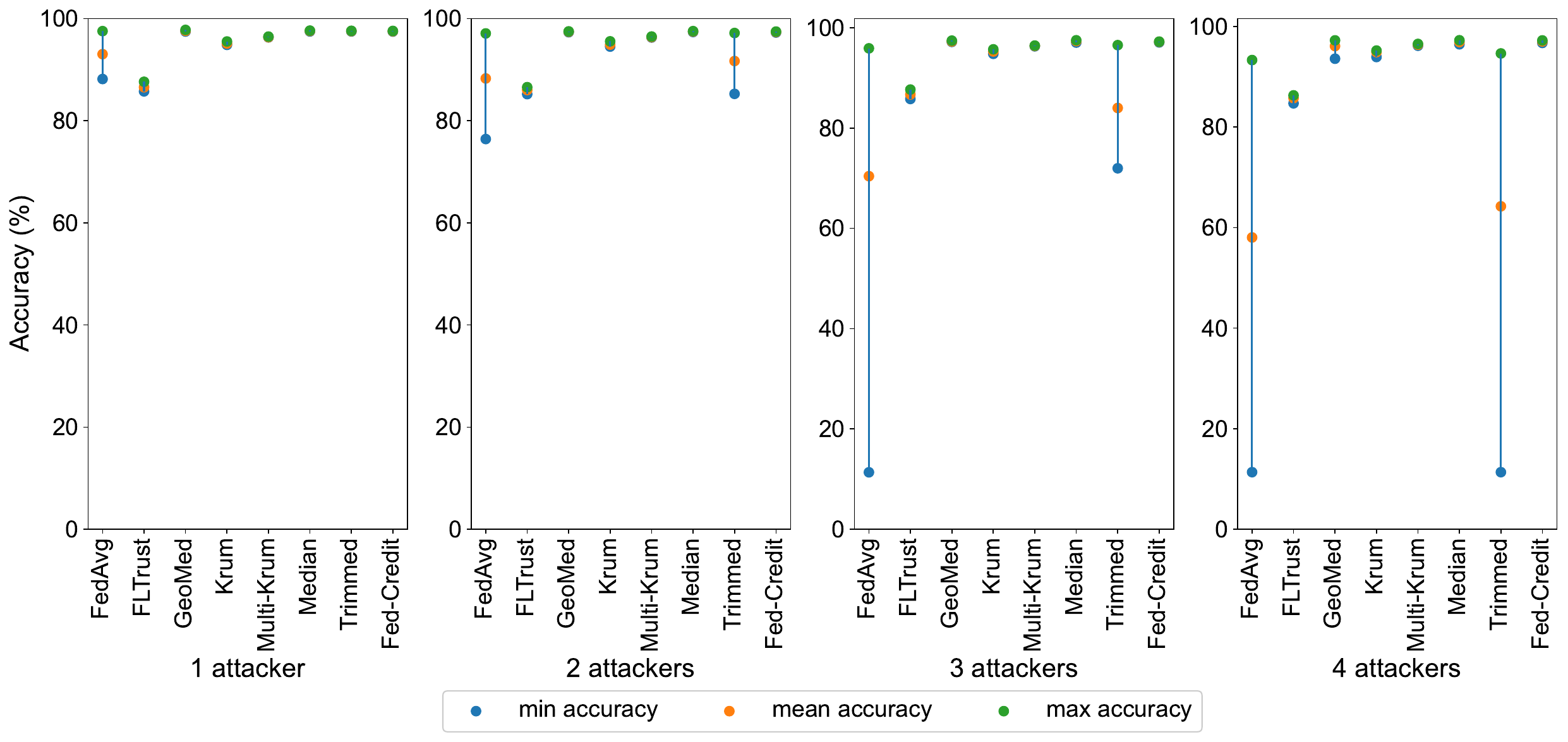}
    \caption{The minimum, mean, maximum accuracy of various aggregation methods with 1,2,3,4 attacker(s) on iid MNIST. {\textit{\rmfamily{Median}}} and {\textit{\rmfamily{\method{}}}} show high accuracy and narrow bias. }
    \label{fig:mnist-iid-acc-attackers}
\end{figure}

\begin{figure}[htbp]
    \centering
    \includegraphics[width=1\textwidth]{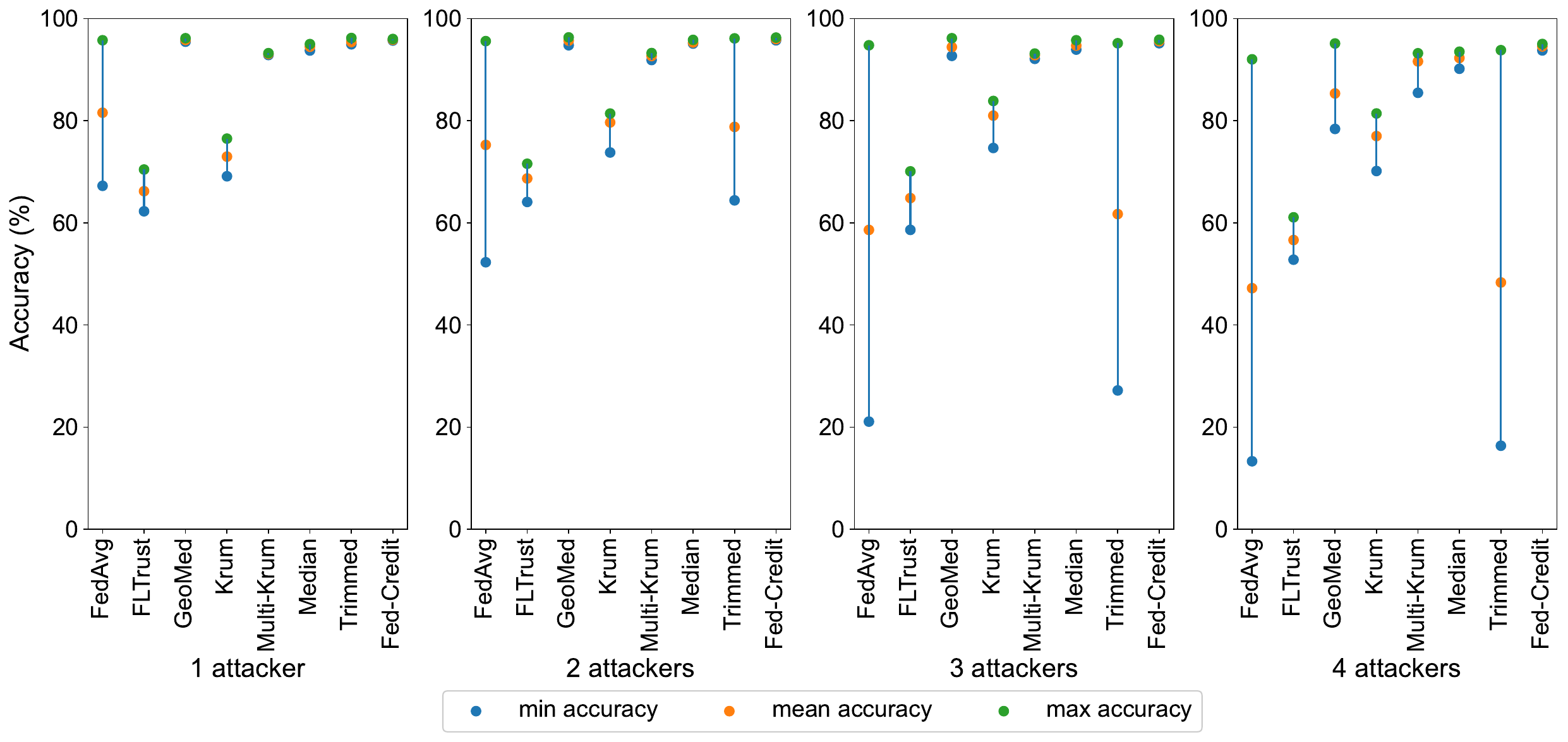}
    \caption{The minimum, mean, maximum accuracy of various aggregation methods with 1,2,3,4 attacker(s) on Non-iid MNIST. {\textit{\rmfamily{\method{}}}} shows high accuracy and narrow bias.}
    \label{fig:mnist-Non-iid-acc-attackers}
\end{figure}

As indicated by the data presented in \Cref{fig:mnist-iid-acc-attackers} \Cref{fig:mnist-Non-iid-acc-attackers}, a clear pattern emerges where an increase in the number of malicious clients corresponds to a noticeable decline in accuracy. Additionally, it is worth noting that the FedAvg and Trimmed algorithms appear to be sensitive to the growing proportion of malicious attackers. This sensitivity can be attributed to the fact that both of these algorithms primarily rely on averaging methods to update the global model. When a malicious client is involved in the computation, it holds the same weight as a benign client, thereby contributing to a degradation in the overall model performance.

In contrast, the results also underscore the superior performance of the \method{} algorithm. Notably, the \method{} algorithm consistently maintains higher accuracy levels and demonstrates fewer instances of extreme variability. This fortifies the assertion that \method{} adeptly preserves both accuracy and stability, even amidst the escalating presence of adversarial entities.

\subsubsection{Impact of Attack Types}\

\begin{figure}[htbp]
    \centering
    \includegraphics[width=1\textwidth]{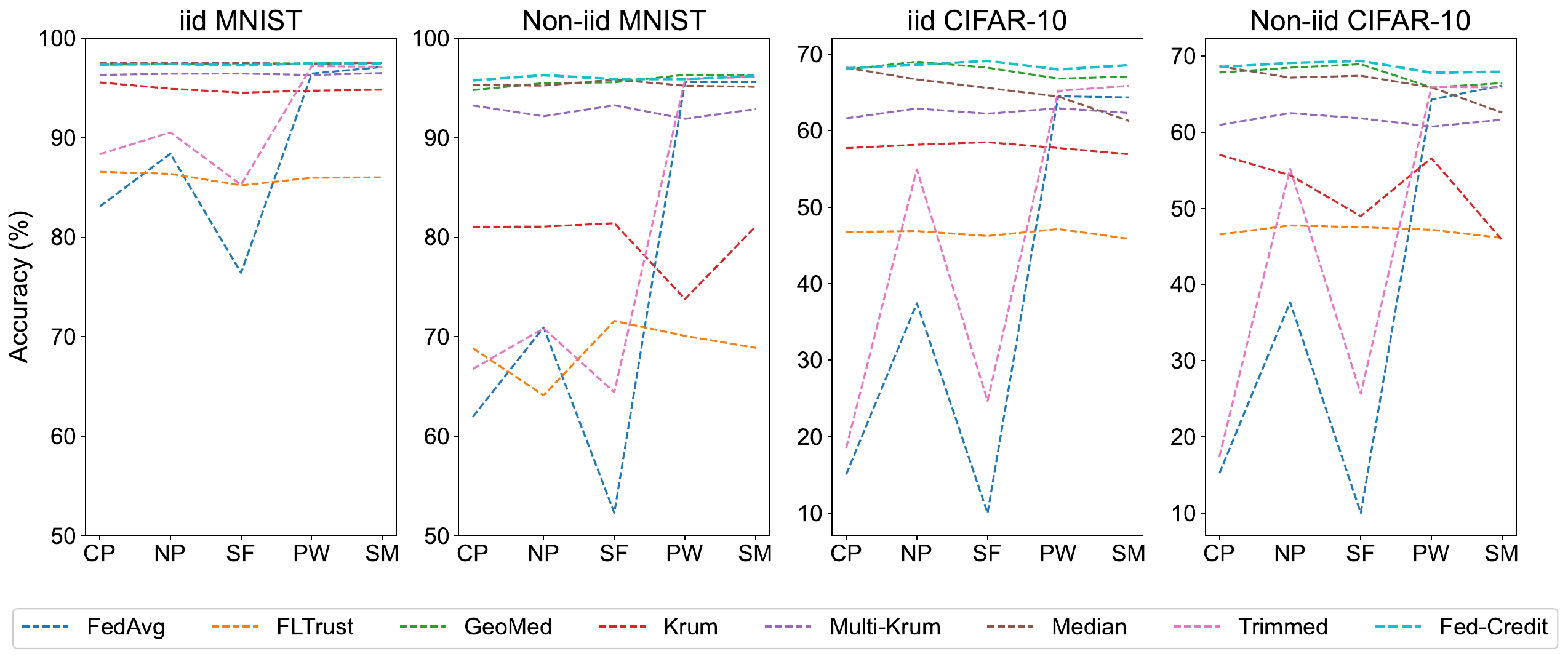}
    \caption{Impact of different attack types on test accuracy for iid and Non-iid datasets. {\textit{\rmfamily{Geomed, Trimmed}}} and {\textit{\rmfamily{\method{}}}} show higher tolerance than other methods.}
    \label{fig:2-attackers}
\end{figure}

It is worth discussing the impact of different attack types on the overall accuracy of the global model. From \Cref{fig:2-attackers}, it is evident that various aggregation approaches show differing levels of effectiveness in countering a range of attack techniques across diverse datasets with distinct distributions. Specifically, when considering the scenario with two attackers, distinct patterns emerge.

For instance, both the FedAvg and Trimmed methods exhibit lower tolerance for Model Poison attacks (CP, NP, SF) compared to Data Poison attacks (PW, SM). On the contrary, algorithms like Krum, Multi-Krum, and FLTrust demonstrate a higher degree of tolerance for multiple attack types. Importantly, these algorithms exhibit sensitivity to only a limited number of attacks, with fluctuations that remain relatively contained compared to FedAvg and Trimmed. The \method{}, Geomed, and Median algorithms consistently perform well, effectively mitigating all types of attacks with higher accuracy compared to alternative methods considered.

\subsubsection{Impact of Data Distribution}\

We conducted an assessment of the model's performance across distinct partitioned datasets. In scenarios where the data partition adheres to the iid principle, an equitable apportionment of each data category to every client was effected. Conversely, in instances characterized by Non-iid data distribution, the \textbf{Dirichlet Distribution} (\( G\sim DP (\alpha, G_{0}) \)) was employed as a means to characterize the prevailing data distribution dynamics.

Prior investigations \cite{non-iid-affect1} \cite{non-iid-affect2} have previously demonstrated the influence of Non-iid datasets on the convergence behavior of models. Our present study, \Cref{fig:mnist-iid-acc-attackers} \Cref{fig:mnist-Non-iid-acc-attackers}, confirmed this view. As the data distribution shifts from iid to Non-iid, the vast majority of methods show a downward trend in accuracy. In line with these antecedent findings, our own experimental endeavor reveals a supplementary facet: that adversarial attacks exhibit heightened efficacy in instances where the underlying dataset distribution is Non-iid. Notably, among the algorithms assessed, namely \method{}, GeoMed, Krum, Multi-Krum, Median, and FLTrust, their predictive accuracy attains a comparable level to that observed under iid dataset conditions when confronted with Non-iid dataset configurations. However, it is noteworthy that both FedAvg and Trimmed algorithms manifest certain challenges in convergence within select scenarios. A case in point involves the application of 3 sign-flipping attackers on the Non-iid CIFAR-10 dataset, where these algorithms nearly regress to a state akin to random conjecture.

\subsubsection{Credibility Trend}\

\begin{figure}[hbp]
    \centering
    \includegraphics[width=1\textwidth]{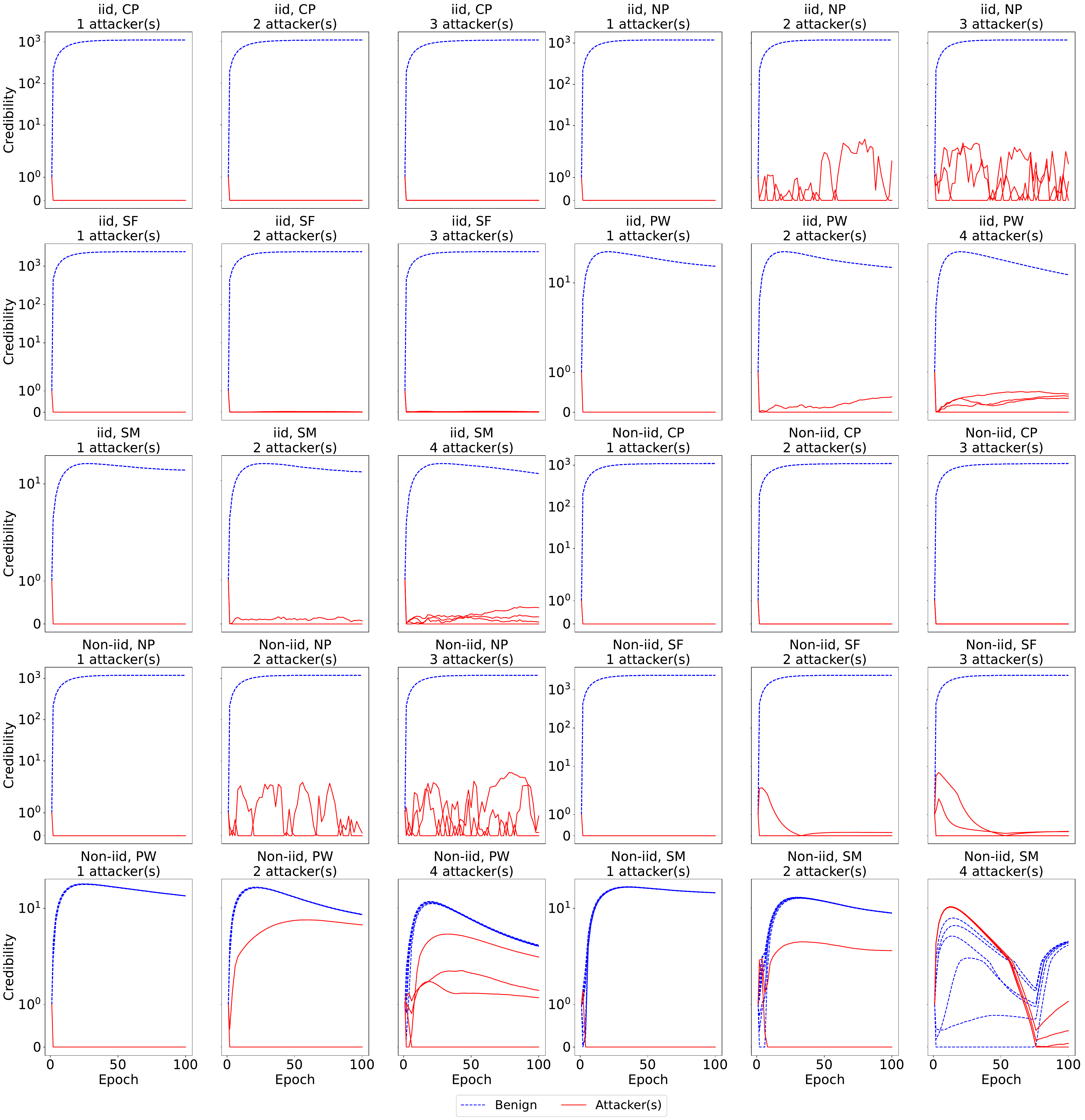}
    \caption{Credibility values of benign clients and attackers with different attack types and varying number of attackers.}
    \label{fig:Credibility}
\end{figure}

Within our \method{} algorithm, a key component that warrants attention is credibility management. This pivotal element profoundly influences the algorithm's operational framework. The graphical representation of credibility values across varying scenarios, employing the MNIST dataset, is concisely depicted in \Cref{fig:Credibility}. This illustration vividly showcases the algorithm's remarkable ability to withstand a diverse array of attacks.

A noteworthy discovery is the consistent trend of credibility values among benign clients, which converged to the same value that is significantly higher than the credibility values that malicious clients achieved. This observation serves as compelling evidence of the \method{} algorithm's effectiveness.

For the group of malicious clients, aside from constant parameter attacks, their credibility values exhibit fluctuations as the number of attackers increases. This phenomenon becomes particularly pronounced when dealing with a Non-iid distributed dataset. It is essential to highlight, however, that this observed fluctuation remains well-contained within an acceptable and manageable scope. Overall, the \method{} algorithm demonstrates a robust and reliable performance, showcasing its resilience across a wide spectrum of challenges.

\section{Related Work}

In this section, we show the current research on poisoning attacks and aggregation rules for defending against attacks.

\subsection{Poisoning Attacks}

According to the poisoning attacks method, poisoning attacks can be classified into data poisoning attacks and model poisoning attacks \cite{attack_classify}.

In the data poisoning attack, the attackers can only inject poison into training data or labels. Therefore, we can divide the data poisoning attack into two categories, \textit{clean label attack and dirty label attack}.

\textbf{Clean label attack:} The untargeted attack \cite{untargeted_attack_1} \cite{untargeted_attack_2} is a form of model poisoning attack. In this attack, malicious clients send arbitrary or counterfeit parameters to the central server with the aim of undermining the performance of the global model or causing it to deviate from its intended behavior. Ali~\etal{} \cite{poison_via_feature_collision} proposed a method that optimizes an equation to create a poison instance resembling a base class instance but embedded in the target class distribution. Dazhong~\etal{} \cite{FedRecAttack} designed FedRecAttack to employ public interactions for approximating the user's feature vector, which an attacker can exploit to train a malicious model. However, the above methods both assume the distribution of the dataset is iid, and if the distribution is Non-iid, the attackers cannot attack via these methods. To address this problem, Jiale~\etal{} \cite{PoisonGAN} utilize a generative adversarial network, called \textit{PoisonGAN}, to generate data similar to other clients and execute attacks with these fake data, in which attackers could execute poisoning attacks without prior knowledge.

\textbf{Dirty lable attack:} Virat~\etal{} \cite{SLF&DLF} introduced that all label flipping can be divided into static label flipping (SLF) and dynamic label flipping (DLF). For instance, an attacker flips the label of "7" to "1" \cite{label_flip_7_1_(1)} \cite{label_flip_7_1_(2)} in SLF. This method has high requirement for prior knowledge which is not inefficient in practical application. To improve efficiency, symmetric flipping \cite{symmetric_attack} and pairwise flipping \cite{pairwise_attack} were introduced to flip each label to other labels. The attack distance-aware attack (ADA) was proposed by Yuwei~\etal{} \cite{ADA} to enhance poisoning attacks by discovering optimal target classes in the feature space.

Model poisoning aims to attack a global model by manipulating malicious clients' local model parameters directly. Li~\etal{} \cite{same_value&sign_flip} use the Same-value vector and Sign-flipping vector to attack the global model. Xie~\etal{} \cite{IPM} proposed Inner Product Manipulation (IPM) which aims to create a negative inner product between the genuine update mean and the aggregation schemes' output, thereby preventing any loss reduction. Wallach~\etal{} designed ALIE to modify the local model parameters carefully based on the assumption that benign updates are expressed by a normal distribution. Xingchen et al. \cite{optimization-based} proposed an optimization-based model poisoning attack, injecting malicious neurons into the neural network's redundant space using the regularization term. However, the primary issue with this approach was the computational complexity of malicious clients needing to compute the Hessian matrix during attack preparation.

\subsection{Defense Rules}

A variety of robust aggregation rules have been proposed. In general, they can be divided into the following three categories.

Distance-based rules aim to detect and reject abnormal local parameters which is uploaded by malicious clients. Blanchard~\etal{} \cite{Krum&Multi-Krum} proposed Krum and Multi-Krum. Krum chooses one update which is the most closest to its neighbors to update the global model, while Multi-Krum computes the mean of multiple updates to update global model. Cao~\etal{} \cite{Sniper} presented Sniper, which constructs a graph based on Euclidean distances between local parameters, to ignore the updates from malicious clients. Wan~\etal{} \cite{MAB-RFL} designed MAB-RFL, which uses graph theory and principal components analysis (PCA) to distinguish honest and malicious in low-dimensional model space.

In performance-based rules, every update from clients will be evaluated with a clean dataset that is stored in server, then the server assigns weights for each update. Cao~\etal{} \cite{DGDA} proposed a Byzantine-robust distributed gradient algorithm that filters out information from malicious clients by computing a noisy gradient with a small clean dataset and only accepting updates based on a pre-defined condition. Zeno \cite{Zeno-2019} uses a small validation set to compute a score for each candidate gradient, considering the estimated loss function descendant and the update magnitude, indicating reliability and performance. Cao~\etal{} \cite{FLTrust} introduced FLTrust, which computes weights by ReLU-clipped cosine similarity between each local update and server update.


Statistics-based algorithms utilize statistical characteristics of updates to update the global parameters. Yin~\etal{} \cite{median&trimmed-mean} proposed Median and Trimmed to exploit the median of updates or the coordinate-wise trimmed mean of local parameters. Xie~\etal{} \cite{Geomed} employed the geometric median, which requires more computational resources, to defend against the attacks. Mhamdi~\etal{} \cite{Bulyan} designed Bulyan, which combines malicious client detection algorithms, such as Multi-Krum, and Trimmed, to filter the updates from malicious clients.

\textbf{\textit{Summary:}} \textit{(1)} Although the current research has good results in defending against some kinds of attacks, few studies have discussed the effectiveness of aggregation rules against multiple attacks. \textit{(2)} For the second category, it's impractical for the server to have a partially clean dataset due to privacy concerns. \textit{(3)} some aggregation rules need to know in advance how many malicious clients there are, which cannot be put into practice. \textit{(4)} high time complexity of one round of interaction for some aggregation rules.

\section{Conclusion}

In this paper, we first explored the practical use of the Federated Learning (FL) algorithm. We then proposed and evaluated a robust FL aggregation method named \method{}. Through extensive experiments on MNIST and CIFAR-10 datasets, we compared \method{} with several other algorithms. Results show that \method{} maintains high accuracy while effectively countering a broad range of attacks. In our future work, we plan to integrate an outlier detection algorithm at the start of \method{} to mitigate extreme local updates and preserve the credibility value system. Additionally, we aim to enhance the generality of \method{} by providing clients with an initial credibility value from previous FL tasks.

\begin{acks}
Acknowledgments goes here...
\end{acks}

\bibliographystyle{ieeetr}
\bibliography{references}


\end{document}